%% file: Draft.tex
\documentclass{article}
\usepackage{spconf,amsmath,graphicx}
\usepackage{xcolor}
\usepackage[ruled,vlined]{algorithm2e}
\usepackage[normalem]{ulem}
\usepackage{amssymb}

\usepackage{tabularx}
\usepackage{multirow}
\usepackage{booktabs}
\usepackage{subcaption}
\usepackage{enumitem}
\usepackage{float}
\usepackage{url}

\newcommand\mycommfont[1]{\footnotesize\ttfamily\textcolor{blue}{#1}}
\SetCommentSty{mycommfont}

\newcommand{\etal}{\textit{et al.}}

\newcommand{\todo}[1]{\textcolor{red}{Complete here}}
\title{Ultra-low bitrate video conferencing using deep image animation}
\name{Goluck Konuko$^{\dagger}$, Giuseppe Valenzise$^{\ddagger}$, St\'ephane Lathuili\`ere$^{\dagger}$}
\address{$^{\dagger}$ LTCI, T\'{e}l\'{e}com Paris, Institut polytechnique de Paris, France\\
$^{\ddagger}$ Universit\'e Paris-Saclay, CNRS, CentraleSup\'elec, Laboratoire des signaux et syst\`emes, France}
\begin{document}
%
\maketitle
\input{abstract}
\setlength{\textfloatsep}{5pt}
\input{intro}

\input{related}

\input{method}

\input{expe}

\input{conclusion}

\vfill\pagebreak

\bibliographystyle{IEEEbib}
\bibliography{strings,refs}

\end{document}

%% file: abstract.tex
\begin{abstract}
In this work we propose a novel deep learning approach for ultra-low bitrate video compression for video conferencing applications. To address the shortcomings of current video compression paradigms when the available bandwidth is extremely limited, we adopt a model-based approach that employs deep neural networks to encode motion information as keypoint displacement and reconstruct the video signal at the decoder side. The overall system is trained in an end-to-end fashion minimizing a reconstruction error on the encoder output.
Objective and subjective quality evaluation experiments demonstrate that the proposed approach provides an average bitrate reduction for the same visual quality of more than 80\% compared to HEVC.  \end{abstract}

%% file: intro.tex
\section{Introduction}
\label{sec:intro}
Video conferencing applications represent a substantial share of Internet video traffic, which has significantly increased in the past months due to the global pandemic. Video conferencing relies heavily of the availability of efficient video compression standards, such as H.264, HEVC~\cite{sullivan2012overview}, and, in the next future, VVC~\cite{bross2019jvet}. Although these video codecs have been optimized and tuned for over 30 years, they are still unable to provide acceptable performance at very low bitrates. In fact, when bandwidth is extremely limited, e.g., due to a congested network or to poor radio coverage, the resulting video quality becomes unacceptable, degrading the video conferencing experience significantly.

A major limitation of the current video compression paradigm comes from employing elementary models of pixel dependencies in order to increase coding efficiency. For instance, the commonly used discrete cosine transform implicitly assumes that pixels are generated by a Gaussian stationary process~\cite{pad2013optimality}. Similarly, spatial and temporal prediction exploit low-level pixel dependencies such as spatial directional smoothness and simple translational motion across time. Furthermore, these tools are often optimized with pixel-based fidelity metrics such as the Mean Squared Error (MSE), which is known to correlate poorly with human perception~\cite{wang2004image}.

In this paper, we consider a very different, \textit{model-based} compression approach for video conferencing that goes beyond simple pixel-based techniques. Specifically, we interpret video frames as points of a low-dimensional manifold living in the higher-dimensional pixel space. We leverage recent advances in image generation and deep generative networks~\cite{goodfellow2015gans}, which have made possible to automatically synthesize images or videos of faces with unprecedented quality and realism~\cite{karras2019style,zakharov2019few, siarohin2019first}. These tools have recently reached a great deal of popularity in computer vision and computer graphics applications, such as motion transfer~\cite{chan2018dance}, creation of video portraits~\cite{kim2018deep}, deep fakes~\cite{yang2019exposing}, etc. However, their potential as a video compression tool is still unexplored. In this work, we describe, for the first time, a video coding pipeline that employs a recently proposed image animation framework~\cite{Siarohin_2019_CVPR,siarohin2019first} to achieve long-term video frame prediction. Our scheme is open loop: we encode the first frame of the video using conventional Intra coding, and transmit keypoints extracted by subsequent frames in the bitstream. At the decoder side, the received keypoints are used to warp the Intra reference frame to reconstruct the video. We also propose and analyze an adaptive Intra frame selection scheme that, in conjunction with varying the quantization of Intra frames, enable to attain different rate-distortion operating points. Our experiments, validated with several video quality metrics and a subjective test campaign, show average bitrate savings compared to HEVC of over 80\%, demonstrating the potential of this approach for ultra-low bitrate video conferencing.

%% file: related.tex
\section{Related Work}
\label{sec:related}

Model-based compression paradigms are not new in data compression. A prominent example comes from speech coding standards, where vocoders~\cite{spanias1994speech} are a broadly used kind of generative models. Unfortunately, generative image and video models turn out to be significantly more complex than in the audio case. Previous work aimed at using generative models for talking heads has failed to provide realistic and appealing video compression performance. The most notable example was the MPEG-4 Visual Objects standard, which enables to embed synthetic objects such as faces or bodies and animate them using a sequence of animation parameters~\cite{Preda2002mpeg4}. Given the resulting animated faces are quite unrealistic, this part of the standard has been rarely used in practice. In this work, we use a recently proposed image animation model which can instead produce highly realistic faces~\cite{siarohin2019first}.

Recently, deep neural networks have been successfully applied to image and video compression~\cite{ma2019image}. Learning-based image compression has been generally cast as the problem of learning deep representations, typically by means of deep auto-encoders, which are optimized in an end-to-end fashion by maximizing a rate-distortion criterion~\cite{balle2018variational, rippel2017real}. 
Learning-based image codecs can compress pictures in a more natural way than conventional codecs~\cite{valenzise2018quality}. Similar ideas have been employed later for the case of video. There, deep learning tools have been mainly used to replace and optimize single stages of the video coding pipeline~\cite{rippel2019learning,wang2019enhancing}. In this work, we consider instead a generative modeling perspective, and propose a coding architecture which departs substantially from that of a conventional hybrid video codec.

Deep generative models have also been recently employed in image/video compression schemes~\cite{agustsson2019generative,kaplanyan2019deepfovea}, to reduce bitrate by hallucinating parts of the video that are outside the region of interest (ROI). In this work, instead, we use deep generative models to synthesize face images, which constitute the main ROI in video communication. ROI-based coding has been previously used for video conferencing~\cite{meddeb2014region}. There, coding gains are obtained by varying the bitrate allocation between the ROI/non-ROI regions. We use instead a completely different warping tool to animate the ROI with a very low bitrate.

%% file: method.tex
\begin{figure*}[t]
	\begin{center}
		\includegraphics[width=0.85\linewidth]{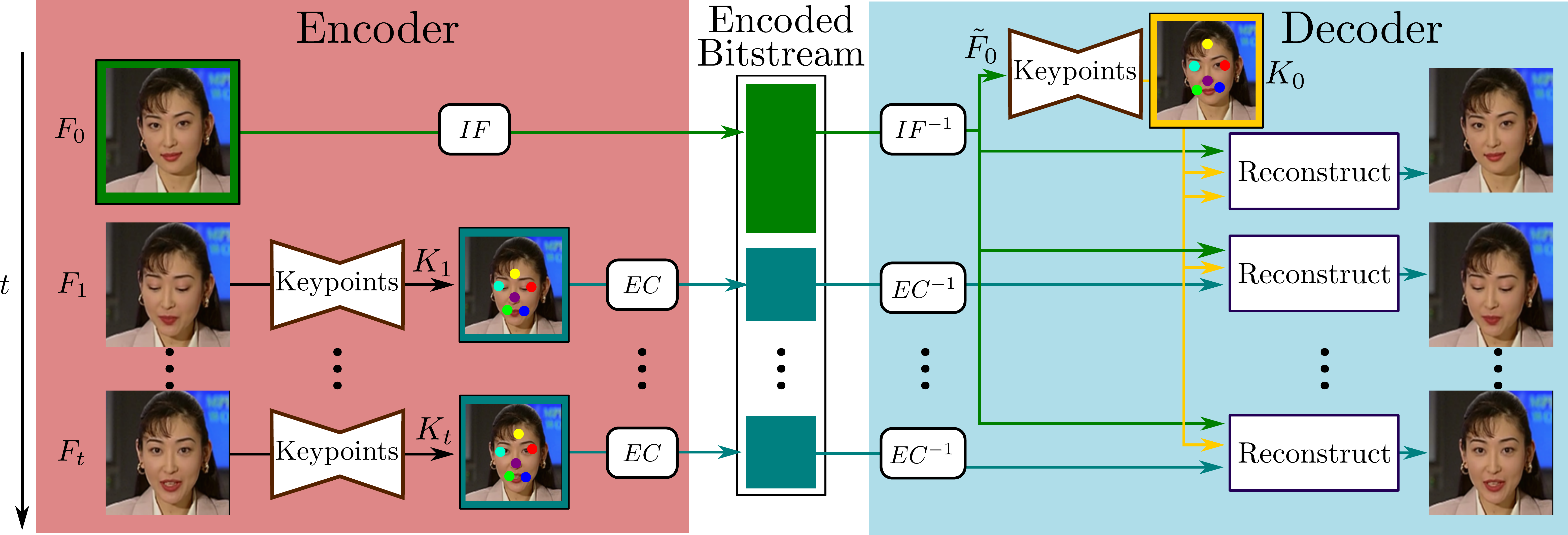}
	\end{center}
    \vspace{-0.5cm}
	\caption{Basic scheme of our proposed baseline codec. The encoder (in red) encodes the first frame as Intra (IF) using a \emph{BPG} codec. The following frames ($t>1$) are encoded as keypoints predicted by a neural network. Keypoints are entropy coded (EC) and transmitted in the bitstream. The decoder (in blue) includes a neural network that combines the reconstructed initial frame $\tilde{F}_0$ and its keypoints $K_0$, with the keypoints $K_t$ at the current time $t$, to reconstruct $F_t$. 
	The overall system is trained in an end-to-end manner via reconstruction loss minimization.}
	\label{fig:pipeline}
\end{figure*}

\section{Proposed coding method}
\label{sec:method}
\subsection{Coding scheme based on image animation}
\label{sec:pipeline}
The overall pipeline of the proposed codec is illustrated in Fig.~\ref{fig:pipeline}. It consists of an Intra frame compression module (which could be any state-of-the-art image codec); a sparse keypoint extractor to code Inter frames; a reconstruction (warping) module to motion compensate and reconstruct the Inter frames; and a binarizer and entropy coder to produce a compressed binary bitstream.  In the following, we provide a walk-through of the main coding steps of the proposed system:

\noindent\textbf{Intra-frame coding.} We first encode the initial frame $F_0$ using the \emph{BPG} image codec, which essentially implements HEVC Intra coding, using a given $QP_0$. The coded frame is sent to the decoder. On the decoder side, the initial frame is decoded using the \emph{BPG} decoding procedure. This decoded initial frame is referred to as $\tilde{F}_0$.

\noindent\textbf{Keypoint prediction.} Following the recent work of Siarohin \etal\cite{siarohin2019first}, we encode motion information via a set of 2D keypoints learned in a self-supervised fashion. The keypoints are predicted by a U-Net architecture~\cite{ronneberger2015u} network that, from its input frame, estimates $M$
heatmaps (one for each keypoint). Each heatmap is then interpreted as detection confidence map and used to compute the expected keypoint locations. In addition, following~\cite{siarohin2019first}, we compute a (symmetric)  $2\times 2$ Jacobian matrix for each keypoint, which encodes the orientation of the face region associated to that keypoint. Each keypoint is then represented by a 5-dimensional vector (2 spatial coordinates, plus the 3 elements to represent the Jacobian matrix). 

\noindent\textbf{Entropy coding.} Keypoints are represented as floating point values with 16 bits precision. We compress these values using an off-the-shelf LZW encoder (\textit{gzip}). Notice that any other entropy coding solution, e.g., a binary arithmetic codec with contexts, could be equally used. The coded motion information, together with the \emph{BPG} compressed Intra frames, form the bitstream that is sent to the decoder. 

\noindent\textbf{Motion compensation and inter-frame reconstruction} At the decoder side, we reconstruct the frame $F_t$ at time $t$ from the decoded initial frame $\tilde{F}_0$ and the displacement of a set of $M$ keypoints, $K_0~\in\mathbb{R}^{5 \times M}$ and $K_t~\in\mathbb{R}^{5 \times M}$, in the frames $\tilde{F}_0$ and $F_t$, respectively. These keypoints are detected for each frame at the encoder side and transmitted in the bitstream, as explained above.   
We then employ the reconstruction network introduced in~\cite{siarohin2019first} to reconstructed $F_t$ from $\tilde{F}_0$ and the keypoints $K_0$ and $K_t$. This reconstruction network is composed of two sub-networks. The first sub-network predicts the optical flow between the frames at times $0$ and $t$ from $\tilde{F}_0$, $K_0$ and $K_t$. This optical flow is then used by the second sub-network to predict $F_t$ from $\tilde{F}_0$. Please refer to~\cite{siarohin2019first} for the details of the reconstruction network architecture. 

The overall system is trained in an end-to-end fashion on a large training set of videos depicting talking faces. We minimize a reconstruction loss between the input and the decoded frames. In this way, we force the keypoint detector network to predict 2D keypoints that describe motion in such a way that the reconstruction network can correctly reconstruct each frame. The reconstruction loss is based on the perceptual loss of Johnson \etal~\cite{johnson2016perceptual} using a VGG-19 network pre-trained on ImageNet. This loss is completed with a GAN loss \cite{mao2017least} and an equivariance loss that enforces that network to be equivariant to random geometric transformations \cite{siarohin2019first}.
Notice that Intra frame coding with \emph{BPG}, as well as \textit{gzip}, are not differentiable, and as a consequence, cannot be included in the training process. While differentiable approaches to estimate entropy have been proposed~\cite{balle2018variational}, in this work we opt for the simpler solution to ignore this step during training. 

\subsection{Adaptive Intra frame selection}
The coding scheme described in the previous section adopts an open GOP structure, i.e., only the first frame is coded as Intra. While this leads to a very high video compression rate, the reconstruction quality might rapidly degrade due to the loss of temporal correlation as frames get farther away from the initial one. This is particularly evident in case of occlusions/disocclusions caused by a person's pose change. 

To overcome this issue, we introduce in the following an adaptive Intra-Refresh scheme using multiple source frames. The procedure is described in Algorithm~1. The proposed multi-source Intra frame selection is parametrized by a threshold parameter $\tau$ that allows quality (and, implicitly, bitrate) adjustment. We employ a buffer $\mathcal{B}$ containing the key frames that are required to reconstruct the input video, together with the corresponding keypoints. At every time step, this buffer is synchronized between the encoder and the decoder. To this end, every frame added to $\mathcal{B}$ by the encoder is sent to the decoder that, in turn, adds the frame to its buffer.

The buffer is initialized with the first frame $F_0$ and the keypoints $K_0$, as described in Section~\ref{sec:pipeline}. Note that the keypoints $K_0$ need not be sent to the decoder since $K_0$ can be re-estimated at the decoder side from $\tilde{F}_0$.
Then, for every frame $F_t$, we apply the following procedure: We estimate the keypoints $K_t$. Then, we identify the best frame in the buffer that can be used to reconstruct the current frame. To this aim, we reconstruct $F_t$ using all the frames $F_b$ and their keypoints $K_b$. We select the buffer frame index $b^*$ that leads to the lowest PSNR. We use PSNR here, rather than perceptual loss, because of its lower computational cost. At this point, two cases can happen: i) if the best source frame yields a PSNR better the threshold $\tau$, we use the frame $F_{b^*}$ as source frame to reconstruct $F_t$. Therefore, the index $b^*$ and the keypoints $K_t$ are sent to the decoder; ii) the PSNR is lower than threshold $\tau$. In this case, none of the frames in the buffer is suitable to reconstruct the current frame and $F_t$ needs to be added to the buffer as a new Intra frame and sent to the decoder.  To this end, we encode the current frame $F_t$ with $QP_0$ used for the first frame. If the resulting PSNR is still above $\tau$, we reduce the QP by one unit, and repeat the procedure till the quality constraint is satisfied. This provides a set of operating rate-distortion curves for the codec. By taking the convex hull of these curves, we can effectively attain different rate-distortion trade-offs. We leave the study of efficient rate-distortion optimization strategies for adaptive Intra frame selection to future work.

\SetKwFor{For}{for}{:}{}%
\SetKwIF{If}{ElseIf}{Else}{if}{:}{elif}{else:}{}%

\begin{algorithm}[t]\label{alg:multi}
\DontPrintSemicolon
  \SetAlgoLined
  \SetKwInOut{Input}{Input}
  \SetKwInOut{Output}{Output}
  \SetKwInOut{return}{Return}
  \Input{Frames: $F_0,\hdots,F_T$ , Threshold: $\tau>0$ }
     $K_0 = \text{\footnotesize\ttfamily Keypoint}(\tilde{F}_0)$ \tcp*{Estimate keypoints}
  $\mathcal{B}=\{(\tilde{F}_0,K_0)\}$ \tcp*{Initialize the buffer}
  $\text{\footnotesize\ttfamily Send}(F_0)$\\
  \For(\mycommfont{\hfill // For every frame}){$t \in \{ 1, ..., T \}$}{
         $K_t = \text{\footnotesize\ttfamily Keypoint}(F_t)$ \tcp*{Estimate keypoints}
      $\tilde{F}^\mathcal{B}_t=\{\text{Reconstruct}(F_b,K_b,K_t); (\tilde{F}_b,K_b) \in \mathcal{B}\}$
      
      $b^*=\underset{\tilde{F}_t^b\in \tilde{F}^\mathcal{B}_t}{\mathrm{argmin}}\left(\mathcal{L}_{rec}(\tilde{F}_t^b,F_t)\right)$\tcp*{Best frame in $\mathcal{B}$}
    \If{$\text{\footnotesize\ttfamily PSNR}(\tilde{F}_t^{b^*},F_t)>\tau$}{$\text{Send}(b^*,K_t)$ \tcp*{Signal the best frame}
               }\Else{ $\text{Set\_QP}(F_t,\tau)$\tcp*{Set QP for IC}
                           $\mathcal{B}\leftarrow \mathcal{B}\cup (\tilde{F}_t,K_t)$ \tcp*{Add to buffer}
           $\text{Send}(F_t)$ \tcp*{Send a new source frame} 
         }
        
}

 \caption{Adaptive Intra frame selection (encoder side)} 

\end{algorithm}

Note that the estimation of $b^*$ could be performed also on the decoder side, since the decoder has access to the synchronized buffer. Re-estimating $b^*$ would indeed avoid the need of sending it, but would require an significant additional decoding complexity. Signaling the best frame index is a good trade-off between compression rate and computational cost.
We implement $\mathcal{B}$ as a first-in, first-out  buffer. When a new source frame is added, the oldest source is popped assuming that reconstruction error increases with time. In all our experiments, we use a buffer of size 5. In our preliminary experiments, we observed that a larger buffer size does not bring significant gain but increase memory requirements.



%% file: expe.tex
\section{EXPERIMENTS AND RESULTS}
\label{sec:pagestyle}

\noindent\textbf{Datasets. } We employ two datasets suited for a video conferencing scenario:

\noindent{\textbullet} \textbf{Voxceleb} is a large audio-visual dataset of human speech of 22,496 videos, extracted from YouTube videos. We follow the pre-processing of \cite{siarohin2019first} and filter out sequences that have resolution lower than 256$\times$256 and resize the remaining videos to 256$\times$256 preserving the aspect ratio. This dataset is split into 12,331 training and 444 test videos. For evaluation, in order to obtain high quality videos, we select the 90 test videos with the highest resolution before resizing.

    \noindent$\bullet$ \textbf{Xiph.org} is a collection of videos we downloaded from Xiph.org~\footnote{\url{https://media.xiph.org/video/derf/}}. This repository includes video sequences widely used in video processing (``News'', ``Akiyo'', ``Salesman'', etc.). We select 16 sequences of talking heads that correspond to the targeted video conferencing scenario. The full list of videos used in the experiments is available on the GitHub page of the paper~\footnote{\url{https://github.com/Goluck-Konuko/dac.git}}. The region of interest is cropped with a resolution of 256$\times$256 with speakers' faces comprising 75\% of the frame. 
    Notice that we use this dataset for testing \textit{only}, while the system is trained on the Voxceleb training set. 


\begin{figure}[t]
\footnotesize
\begin{minipage}[c]{0.24\textwidth}
 \includegraphics[width=0.99\textwidth]{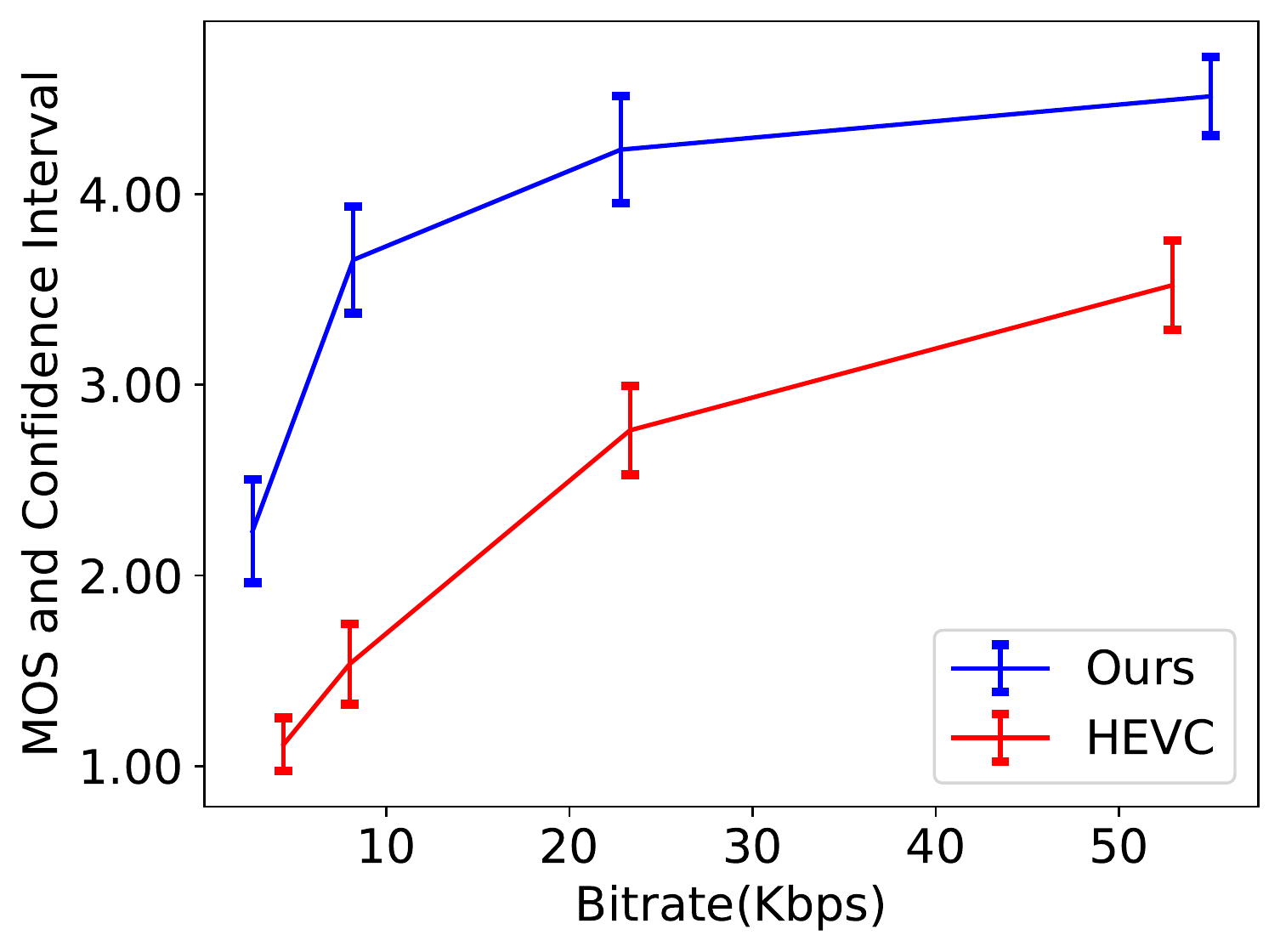}
\end{minipage}
\hspace{6pt}
\begin{minipage}[c]{0.22\textwidth} 
\footnotesize
\begin{tabular}{@{}l@{\hspace{3pt}}c@{\hspace{9pt}}cc@{}}
\toprule
&\textbf{PCC}&\textbf{SROCC}\\
\midrule
\textbf{PSNR}& 0.92/0.95& 0.77/0.92\\
\textbf{SSIM}& 0.91/0.74& 0.80/0.67\\
\textbf{MS-SSIM}& 0.94/0.95 & 0.84/0.95\\
\textbf{VIF} & 0.38/0.75 & 0.28/0.75\\
\textbf{VMAF}& 0.94/0.96 & 0.80/0.89\\
  \bottomrule
\end{tabular}
\end{minipage}
\vspace{-0.4cm}
\caption{User evaluation: Mean Opinion Score (MOS) for HEVC and our codec (average over all sequences). We study correlation between MOS and each metric for videos compressed with our approach/HEVC.}
\label{fig:RD_HEVC}
\end{figure}

\noindent\textbf{Comparison with HEVC.}
We compare our approach with the standard HM 16 (HEVC) low delay configuration\footnote{{\url{https://github.com/listenlink/HM}}} by performing subjective test on 10 videos (8 from Voxceleb and 2 from Xiph.org) using Amazon Mechanical Turk (AMT). Each sequence is encoded using 8 different bite-rate configurations (4 with HEVC, 4 with proposed method) ranging from 5Kbps to 55Kbps. The different points in the curve are obtained by changing $QP_0$ in BPG, as well as $\tau$. We implement a simple Double Stimulus Impairment Scale (DSIS) test \cite{BT500} interface in AMT. Users are invited to follow a brief training on a sequence not used for test. The extreme quality levels (``very annoying'', ``imperceptible'') are used as gold units to check the reliability of voters. We collected 30 subjective evaluations per stimulus. After this screening, no further outliers were found. The Mean Opinion Scores (MOS) with 95\% confidence intervals are shown in Fig.~\ref{fig:RD_HEVC}-left. It shows that our approach clearly outperforms HEVC (average BD-MOS = 1.68, BD-rate = -82.35\%). In Fig~\ref{fig:RD_HEVC}-right, we report the Pearson Correlation Coefficient (PCC) and the Spearman  Rank-Order  Correlation  Coefficient (SROCC) between MOS and five commonly used quality metrics, for our codec and HEVC. We observe that, except VIF, these metrics correlate well with human judgment, at least on the tested data. Based on these preliminary results, we proceed with an extensive performance evaluation of the proposed method using these quality metrics.
In Table~\ref{tab:Comp-HEVC}, we report the average Bjontengaard-Delta performance for VoxCeleb test and Xiph.org images. The gains are significant (over 80\% BD-rate savings) on the two datasets. 

\begin{figure}
  \centering
\bgroup
\def\arraystretch{0.1}
\setlength\tabcolsep{1.1pt}
\begin{tabular}{ccccc}
\footnotesize Original \footnotesize  & \footnotesize HEVC: \textcolor{red}{17.1Kbps}
& \footnotesize Ours: \textcolor{blue}{15.7Kbps}& \footnotesize  Ours: \textcolor{blue}{2.3Kbps}  \\
\includegraphics[width=0.116\textwidth]{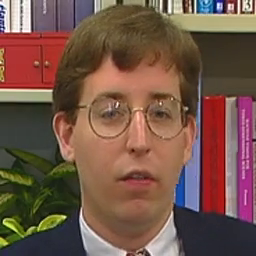}&
\includegraphics[width=0.116\textwidth]{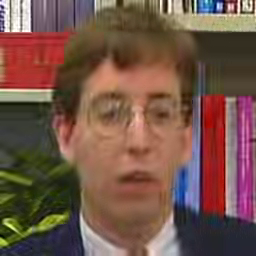}&
\includegraphics[width=0.116\textwidth]{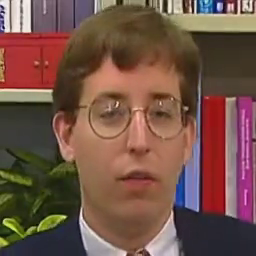}&
\includegraphics[width=0.116\textwidth]{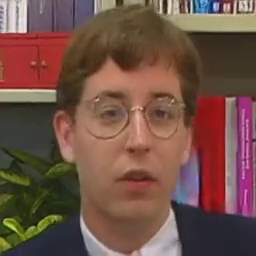}\\
\\
\includegraphics[trim=60 100 60 95,clip,width=0.116\textwidth]{figures/ablation_study/mos_seq/low_br/org_3.png}
&\includegraphics[trim=60 100 60 95,clip,width=0.116\textwidth]{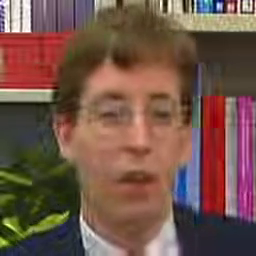}
&\includegraphics[trim=60 100 60 95,clip,width=0.116\textwidth]{figures/ablation_study/mos_seq/low_br/dac_3.png}
&\includegraphics[trim=60 100 60 95,clip,width=0.116\textwidth]{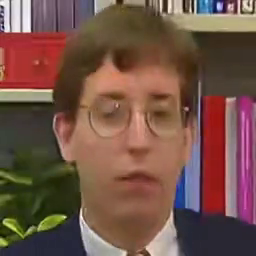}

\\
\end{tabular}
\egroup
\vspace{-0.3cm}
\caption{Qualitative evaluation against HEVC on \emph{Deadline} sequence in a low bitrate setting.}
\label{fig:quali}
\end{figure}

Finally, we provide a qualitative comparison in Fig.~\ref{fig:quali} on a sequence from Xiph.org. In this example, we observe that at a similar bitrate, our approach produces much fewer artifacts. The difference is clearly visible in the eye region. Even at a much lower bitrate (2.3 Kbps) -- one lower than the minimum rate achievable by HEVC -- our approach generates better quality images. 

\begin{table}[t]
\begin{center}
\caption{Bjontengaard-Delta Performance over HEVC}
\vspace{-0.3cm}
\footnotesize
\begin{tabular}{lccc}
\toprule
&\textbf{VoxCeleb}&\textbf{Xiph.org}\\
&{\footnotesize BD quality~/~BD rate}&\footnotesize BD quality~/~BD rate\\
\midrule
\textbf{PSNR}& 2.88~/~-65.50&3.14~/~-72.44\\
\textbf{SSIM}&  ~0.122/~-83.96 &  ~0.02/~-65.79 \\
\textbf{MS-SSIM}&0.070~/~-83.60 & 0.075~/~-86.41\\
\textbf{VIF} & 0.027~/~-72.29 &0.021~/~-68.02\\
\textbf{VMAF}&  37.43~/~-82.29 & 31.04~/~-83.44\\
  \bottomrule
\end{tabular}
\label{tab:Comp-HEVC}
\end{center}
\vspace{-0.5cm}
\end{table}




%% file: conclusion.tex
\section{Conclusions}
\label{sec:conclusion}

In this paper, we propose the first video conferencing codec that employs a deep generative frame animation scheme and drastically improve coding performance at ultra-low bitrate. In our approach, we encode  the  initial frame using a state-of-the-art image codec. Motion information is then encoded via moving keypoints predicted by a deep neural network. We propose an adaptive Intra frame selection mechanism to improve reconstruction quality over long sequences. Our experiments show that our approach outperforms HEVC with a large margin. As future work, we plan to extend our model by including rate-distortion optimization, handling high-resolution videos, and addressing more complex scenarios with multiple speakers in the same video.